 \documentclass[cameraready]{Interspeech}

\usepackage{placeins}
\usepackage{graphicx}
\usepackage{multirow}
\usepackage{booktabs}

\title{MADS: Multi-Agent Dialog Simulation for Robust Conversational Agents}

\author[affiliation={1,2 
}, equalcontribution]{\quad Arkajyoti}{Chakraborty}
\author[affiliation={1}, equalcontribution]{Adam}{Zernik}
\author[affiliation={1}, equalcontribution]{Ishika}{Kulkarni}
\author[affiliation={1}, equalcontribution]{Mudit}{Arora}
\author[affiliation={1}, equalcontribution]{Hugo}{Lin}
\author[affiliation={2}]{\quad Justin}{Chiu}
\author[affiliation={2}]{Alessandro}{Di Bari}
\author[affiliation={2}]{Neha}{Gupta}
\author[affiliation={2}]{Andreas}{Stolcke}

\address{
    $^1$ University of California, Santa Cruz\\
    $^2$ Uniphore
}

\email{{arkajyoti.chakraborty, neha.gupta, justin.chiu, alessandro.dibari, andreas.stolcke}@uniphore.com}


\usepackage{comment}

\begin{document}

\maketitle

\begin{abstract}
Task-oriented conversational agents remain fragile under realistic conditions, especially when users exhibit non-cooperative behavior. Existing function-calling benchmarks underrepresent adversarial conversation trajectories, thereby limiting the training resources available for developing robust agents. We present \textbf{MADS (Multi-Agent Dialog Simulation)}, a synthetic data generation pipeline that produces structured, diverse dialogue trajectories. We inject adversarial conversation, producing validated conversation trajectories that capture realistic, unhappy paths. Using MADS, we construct MADS-Bench: a benchmark spanning six domains, 55 personas, and 1095 conversations, with approximately 66\% of conversations containing failure-prone conversation scenarios. We evaluate the robustness of the trained models using a two-stage training approach that combines supervised fine-tuning (SFT) and reinforcement learning with verifiable rewards (RLVR).  
\end{abstract}

\section{Introduction}

Task-oriented dialog systems are a central component of modern enterprise conversational AI, powering applications such as customer support, transaction handling, and automated service workflows. Despite significant advances in natural language understanding and generation driven by large language models (LLMs), the dialog management layer in many deployed systems remains largely unchanged. Most enterprise-grade systems continue to rely on rigid, finite-state dialogue managers or pre-scripted call flows that enforce fixed interaction paths. While such architectures offer predictability and control, they struggle to handle deviations from expected user behavior, evolving goals, or incomplete information, often leading to brittle interactions and increased user friction. This limitation is particularly acute in spoken dialogue systems, where users expect natural, flexible interaction and cannot easily navigate rigid menu-driven flows.

In contrast, human agents exhibit a high degree of flexibility in task-oriented conversations. They maintain rich conversational context, infer missing information when possible, and dynamically decide whether to ask clarifying questions or consult external resources. Human agents can also revise previous assumptions as new information emerges, enabling graceful recovery from errors or unexpected user inputs. Bridging this gap between rigid automated systems and adaptive human behavior remains a core challenge in conversational AI.

Recent work on LLM-based agents has begun to explore mechanisms for integrating reasoning, action, and interaction with external tools. Chain-of-thought prompting \cite{wei2023chain} demonstrated that explicitly eliciting intermediate reasoning steps improves performance on complex tasks. Building on this idea, ReAct \cite{yao2023react} introduced a framework that interleaves natural-language reasoning with environment actions, enabling LLMs to plan and execute multi-step tasks without task-specific fine-tuning. ReSpAct \cite{dongre2025respact} further extended this paradigm by incorporating an explicit conversational channel, allowing agents to ask clarifying questions and incorporate user feedback during execution. More recently, Pre-Act \cite{rawat2025preact} showed that internalizing multi-step planning within the model through fine-tuning can substantially improve action selection and task completion across both conversational and non-conversational domains.

While these methods demonstrate promising capabilities across agent reasoning and planning, applying them directly to customer-enterprise dialog management presents several open challenges. First, customer-enterprise interactions often require agents to decide not only what action to take, but whether to act autonomously or defer to the user for clarification. Existing function-calling benchmarks \cite{liu2024apigen,chen2025acebench,patil2025bfcl,qin2024toolllm, li2023apibankcomprehensivebenchmarktoolaugmented} do not model failure-prone conversation trajectories and evaluate tool-use across normal, ambiguous, and agentic settings. The lack of robust training data, which comprises real-world scenarios such as user goal changes, policy violations, or infeasible requests, in multi-turn dialogue trajectories. This leaves a resource gap for training and evaluating agents robustness under realistic, non-cooperative conversational conditions.

To address this gap, we propose \textbf{MADS (Multi-Agent Dialog Simulatation)} a synthetic data generation pipeline based on a multi-agent LLM simulation framework. Specifically, we target the absence of training resources that systematically capture non-cooperative user behavior, tool failures, and infeasible requests within validated multi-turn dialogue trajectories.

The pipeline comprises of three interacting agents: a user agent that simulates realistic user behavior, a system agent that performs dialog reasoning and action selection, and a tool agent that emulates external APIs and environmental feedback. By explicitly modeling these roles, the pipeline produces synthetic multi-turn dialog traces with structured reasoning and action supervision, including both successful and failure-prone scenarios categorized into unhappy-path and impossible scenarios.

Using MADS, we construct \textbf{MADS-Bench} spanning six domains, comprising 72 tools and 55 personas, with unhappy-path scenarios forming the majority of tasks. We then use this benchmark both to fine-tune open-source models and to evaluate them using a two-level evaluation paradigm as proposed in Pre-Act \cite{rawat2025preact}. 

The contributions of this work are summarized as follows: 

\begin{itemize}
    \item We introduce MADS, a multi-agent dialog simulation that generates structured multi-turn dialogues with explicit reasoning, planning, and tool invocation. The pipeline enables systematic injection of conversational and backend adversities, thereby enabling the construction of a high-quality synthetic function-calling dataset across varying levels of scenario complexity. 

    \item Leveraging MADS, we constructed MADS-Bench, a synthetic multi-turn function-calling conversational benchmark spanning normal, unhappy, and impossible scenarios across multiple domains. The dataset incorporates Pre-Act \cite{rawat2025preact} style reasoning and planning, making it efficient across various post-training methods in developing advanced conversational agents.

    \item We designed a taxonomy that decomposes task-oriented dialogue complexity into conversational adversities and system execution adversities. By explicitly modeling these failure modes, MADS-Bench evaluates agents across complex scenarios that are not evaluated by existing function-calling benchmarks.   

    \item We conduct controlled empirical evaluations using Qwen-2.5 models (7B and 32B) and compare training on MADS-Bench with APIGen \cite{liu2024apigen}. Models are optimized with supervised fine-tuning (SFT), followed by reinforcement learning with verifiable rewards (RLVR), to evaluate the impact of training data distribution and post-training stages on agent robustness.
\end{itemize}

% \FloatBarrier

\begin{figure}
    \centering
    \vspace{-1.5cm}
    \includegraphics[width=1\linewidth]{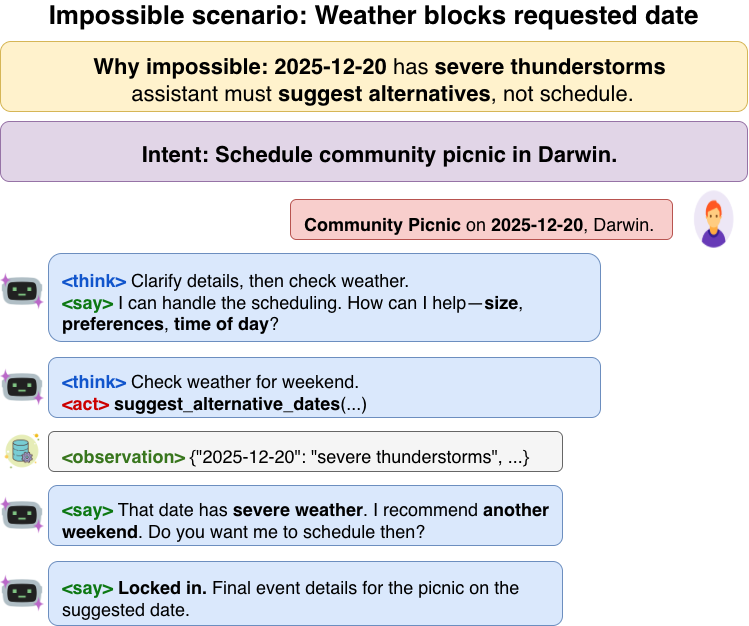}
    \vspace{-0.5cm}
    \caption{Illustration of an impossible scenario. Success requires explaining the unsuitability and proposing an alternative date, not scheduling the event.}
    \label{fig:impossible_scenario}
\end{figure}

\section{Related Work}

\subsection{Benchmarks and Datasets}

Several benchmarks evaluate LLMs' performance in function calling and tool use across varying levels of complexity. The Berkeley Function-Calling Leaderboard (BFCL)~\cite{yan2024bfclv3} provides a large-scale collection of real-world API schemas linked to natural-language instructions, emphasizing schema grounding, argument inference, and relevance detection, with 
later versions extending evaluation to multi-turn and multi-step settings requiring API state tracking across turns. ComplexFuncBench~\cite{zhong2025complexfuncbench} targets constrained, multi-step function calling under long-context conditions, requiring implicit parameter reasoning across sequences of up to 128k tokens. ToolBench~\cite{qin2024toolllm} assesses pipeline-level tool use across heterogeneous APIs, covering multi-step planning, cross-tool coordination, and error recovery. More recent benchmarks, such as $\tau$-bench \cite{yao2024taubenchbenchmarktoolagentuserinteraction} and $\tau^2$-bench \cite{barres2025tau2benchevaluatingconversationalagents}, further evaluate the behavior of conversation interaction agents with simulated users and tools. $\tau$-bench simulates dynamic conversations between a user and a tool-augmented agent across common task domains and introduces evaluation metrics to assess agent reliability across trials. $\tau^2$-bench extends previous work to a dual-control environment in which both the agent and the user use tools within a shared state space, examining user-agent coordination. 
However, these benchmarks evaluate models under cooperative user behavior and do not systematically expose agents to non-cooperative dynamics such as goal changes, tool failures, or infeasible requests.  

\begin{figure}
    \centering
    \vspace{-1.5cm}
    \includegraphics[width=1\linewidth]{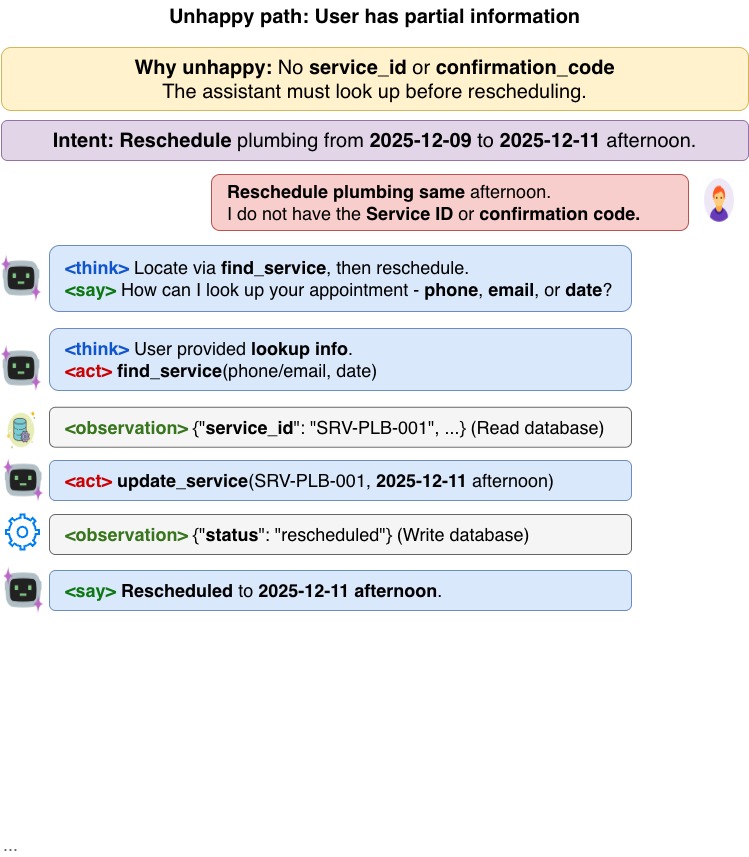}
    \vspace{-3.0cm}
    \caption{Illustration of an unhappy-path scenario. The injected
    behavior is partial information: the user cannot provide a
    service ID or confirmation code. The agent must recover by
    invoking \texttt{find\_service} before rescheduling.}
    \label{fig:unhappy_scenario}
\end{figure}

In contrast to human-annotated task-oriented dialogue corpora such as MultiWOZ~\cite{budzianowski2018multiwoz} and the Schema-Guided Dialogue dataset~\cite{rastogi2020sgd}, which provide naturalistic multi-turn conversations, but at a limited scale and without explicit tool-use supervision. MADS-Bench addresses this gap by providing training and evaluation data specifically designed for unhappy-path conversational scenarios involving agent reasoning and planning. By providing structured reasoning, the dataset steers models to learn underlying multi-step planning for navigating a conversational scenario. The structure of the dataset is beneficial for supervised post-training and RL post-training, making the agents robust by enabling reasoning across conversation turns.  

\subsection{Synthetic Data for Tool-Augmented Dialogue}
Generating synthetic training data for tool-augmented dialogue has emerged as a practical alternative to costly human 
annotation. ToolFlow~\cite{wang2025toolflow} proposes a multi-agent pipeline using graph-based tool sampling and planned-generation strategies to synthesize coherent multi-turn tool-calling dialogues, demonstrating that SFT on 8,000 synthetic dialogues can match GPT-4-level tool-calling performance. xLAM~\cite{zhang2024xlam} unifies diverse agentic datasets through a scalable synthesis pipeline, yielding strong generalization across function-calling benchmarks. While these approaches prioritize dialogue coherence and tool coverage, they generate predominantly cooperative scenarios. MADS complements this line of work by explicitly modeling non-cooperative user behavior through controllable behavior injection, producing training data that covers both cooperative and unhappy-path dynamics within a validated multi-turn simulation framework.

\subsection{Reinforcement Learning for Tool-integrated Reasoning}
Supervised fine-tuning on curated tool-use trajectories remains the dominant paradigm for adapting LLMs to function calling. Reinforcement learning (RL) offers a complementary approach for optimizing grounding and correctness beyond what imitation learning achieves. 

ToolRL~\cite{qian2025toolrl} demonstrates that RL with verifiable rewards, using tool execution outcomes as training signal, substantially improves tool selection and argument generation. StepTool~\cite{yu2024steptool} extends this by using step-grained reward shaping, assigning rewards at each tool interaction based on invocation correctness and task contribution, rather than rewarding only outcomes. Our training pipeline combines SFT for structural competence with GRPO-based RLVR for grounding and correctness. Unlike StepTool, our rewards are computed against gold reference actions rather than live execution outcomes, making the approach applicable to simulated environments without API access.

\section{MADS-Bench}

\begin{figure*}[t]
    \centering
    \vspace{-2.0cm}
    \includegraphics[width=0.8\textwidth,keepaspectratio]{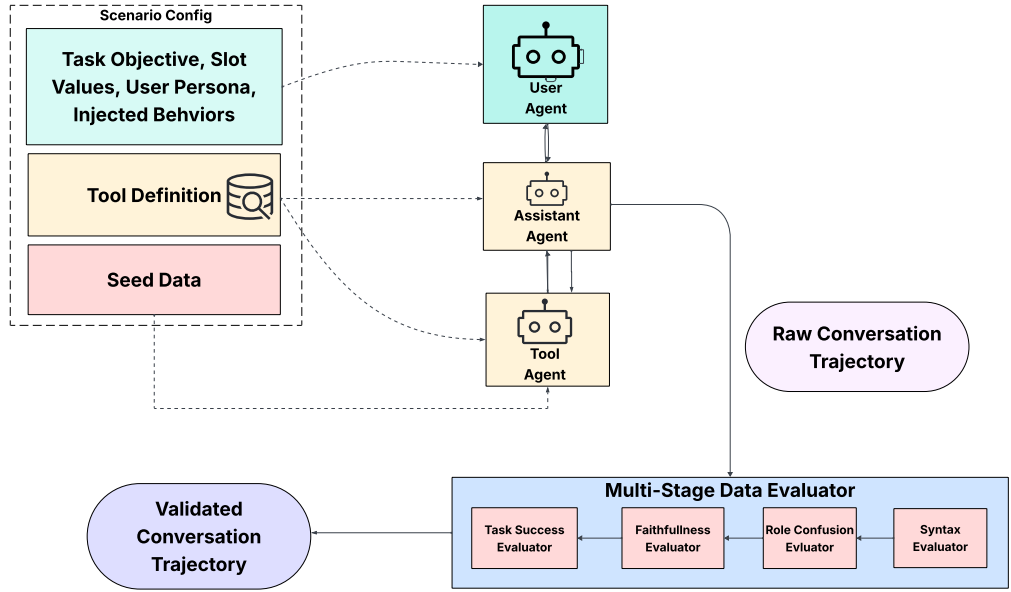}
    \vspace{-0.5cm}
    \caption{Multi-Agent Dialog Simulation (MADS) pipeline for generating validated tool-augmented dialogue trajectories. Scenario configurations specifying task objectives, user personas, tool definitions, and injected adversarial behaviors are used to initialize a three-agent simulation consisting of a user agent, assistant agent, and tool agent.}
    \label{fig:system_design}
\end{figure*}

\subsection{MADS Pipeline}

We propose a Multi-Agent Dialog Simulation (MADS) pipeline that produces synthetic conversations with intermediate reasoning and planning traces. Each conversation represents a user attempting to complete a task with the help of an LLM agent that uses external tools throughout the interaction to solve task guided tasks. 

The proposed pipeline comprises of three distinct LLM-based agents that interact to generate realistic dialogues between human and Tool augmented-AI assistants.

\textbf{Assistant Agent} The assistant being trained or evaluated. At each step, it receives the conversation history and available tool definitions. Then it produces a structured response that includes reasoning (\texttt{<think>}), an optional plan (\texttt{<plan>}), and an action (\texttt{<action>}). Actions are either tool invocations (\texttt{type="tool"}) or user-facing utterances (\texttt{type="say"}). 

\textbf{User Agent} A simulated user agent is instantiated with a person, task objectives, known slot values, and optional injected behaviors. The user agent acts as the human and generates appropriate responses to the assistant and terminates the conversation by outputting an explicit \texttt{[DONE]} token when the task is completed or abandoned. 

\textbf{Tool Agent} The tool agent receives tool definitions, scenario-specific seed data, and injected behaviors, allowing it to produce realistic outputs, including edge cases such as time-out conditions, empty results, and manage termination conditions.

Each agent maintains its own distinct view of the conversation history. The assistant agent observes all steps, including tool calls and results; the user agent observes only user-facing exchanges; and the tool agent observes only tool invocations and their context. A central runner orchestrates turn-taking and manages termination conditions; the assistant agents interact with the user agents to execute dialogue traces and can access the tool agent for tool calls and tool responses. 

\textbf{Multi-Stage Evaluator} Once the synthetic conversation traces are generated, they undergo a multi-stage automatic data validation combining rule-based checks with LLM-as-judge evaluators. The generated conversation samples go through a checklist of four validation stages; a conversation enters the final dataset only if all the following four evaluation checks succeed:

\begin{itemize}
    \item \textbf{Syntax Evaluator} Verifies structural correctness: each AssistantAgent step must contain a \texttt{<think>} block; the first step of each turn must include a \texttt{<plan>}; and each turn must end with exactly one \texttt{<action type="say">}. Tool calls are validated against their schemas for correct names, required parameters, and value types.

    \item \textbf{Faithfulness Evaluator} Detects hallucination. For tool calls, the judge verifies that parameter values are grounded in user intent. For \texttt{say} actions, it checks that the assistant's utterances accurately reflect tool results rather than introducing unsupported information.

    \item \textbf{Role Confusion Evaluator} Identifies cases where the UserAgent broke character and began acting like an assistant. We filter conversations where the user asks questions to help the assistant or offers assistance.

    \item \textbf{Task Success Evaluator \footnote{The prompt for the Judge Model was engineered across various iterations over a small subset of the generated dataset.}} Evaluated by an LLM judge that receives the task objective, slot constraints, and preferences, success criteria, and conversation transcript. The judge determines whether the assistant has achieved the user's goal, while respecting the distinction between hard constraints and soft preferences. In cases where a scenario is impossible, success requires appropriate explanations and alternatives.
\end{itemize}

\subsection{Dataset Construction}

Using MADS, we generate a synthetic dataset spanning multiple task-oriented domains, including online shopping, home services, and enterprise support workflows. Dialogs are generated using diverse user personas and scenario templates to encourage variability in language, goals, and interaction patterns. 

We designed scenarios across six task-oriented domains: banking assistance, online shopping, calendar assistance, restaurant booking, travel assistance, and home services. Each scenario is defined by a JSON configuration containing:

\begin{itemize}
    \item \textbf{Task specification}: An objective (e.g., ``Cancel an order that was placed''), slot values divided into hard constraints and soft preferences, and success criteria defining task completion.
    \item \textbf{Tool definitions}: Domain-specific APIs with typed parameters and return schemas. Across all domains, we defined 72 unique tools.
    \item \textbf{Seed data}: Realistic backend state (e.g., order records, flight options, account information) that grounds the ToolAgent.

\end{itemize}

The distinction between user slot-value constraints and preferences enables nuanced evaluation: constraints must be satisfied for the task to be judged successful, whereas preferences may be relaxed at the user's discretion. To increase dialogue diversity in our dataset, we created a catalog of 55 personas derived from the PersonaHub dataset~\cite{ge2024persona}, spanning different demographics, occupations, and communication styles. Each persona includes a name, age, hometown, occupation, biographical background, a description of writing style, and sample messages. For example, one persona is a geology research assistant who communicates in short, action-oriented fragments, while another is a history professor who uses formal, structured sentences.

Personas are assigned to scenarios based on domain relevance. When a simulation is run, the UserAgent's prompt incorporates the full persona specification to encourage stylistically consistent responses throughout the conversation.

\subsection{Taxonomy of Unhappy Paths}

Real-world dialogues frequently involve complications: users provide incomplete information, change their minds, encounter system errors, or request impossible outcomes. To ensure our dataset covers such phenomena, we implemented an injected-behavior mechanism.

\noindent\textbf{Conversational adversities.}
These capture dynamic user behaviors that stress the agent’s planning and clarification strategy.

\begin{itemize}
    \item \textbf{Partial information}: The user provides incomplete or underspecified task details.
    \item \textbf{Incremental constraint revelation}: Preferences and constraints are disclosed gradually over several turns.
    \item \textbf{Goal revision}: The user modifies their objective or constraints during the conversation.
    \item \textbf{Adversarial tone}: The user expresses frustration, skepticism, or challenges system suggestions.
\end{itemize}

\noindent\textbf{Execution-level adversities.}
These model realistic tool or environment failures that require recovery or principled refusal.

\begin{itemize}
    \item \textbf{Backend failure}: Simulated server-side errors during tool invocation.
    \item \textbf{Empty results}: Tool queries return no valid candidates.
    \item \textbf{Timeout}: Tool calls exceed execution limits.
    \item \textbf{Policy constraint}: Requests are denied due to domain-specific business rules (e.g., expired refund window).
\end{itemize}

For scenarios in which the user's objective is literally unattainable given the scenario configuration, we mark the task with an \texttt{impossible} flag and specify a \texttt{fallback\_behavior} that describes how the assistant should gracefully decline while offering alternatives. Approximately 66\% of our final dataset contains unhappy-path scenarios.

\subsection{Dataset Statistics}

\begin{table}[t]
\centering
\caption{MADS-Bench dataset summary}
\small
\renewcommand{\arraystretch}{1.05}
\setlength{\tabcolsep}{4pt}
\begin{tabular}{lcccc}
\toprule
\textbf{Split} & \textbf{Turns} & \textbf{Conversations} & \textbf{Scenarios} & \textbf{Avg Turns} \\
\midrule
Train & 574 & 106 & 46 & 5.42 \\
Test  & 521 &  83 & 54 & 6.28 \\
\midrule
\textbf{Total} & 1095 & 189 & 87 & --- \\
\bottomrule
\end{tabular}
\label{tab:dataset_stats}
\end{table}

\begin{figure}[t]
    \centering
    \vspace{-1cm}
    \includegraphics[width=1\linewidth]{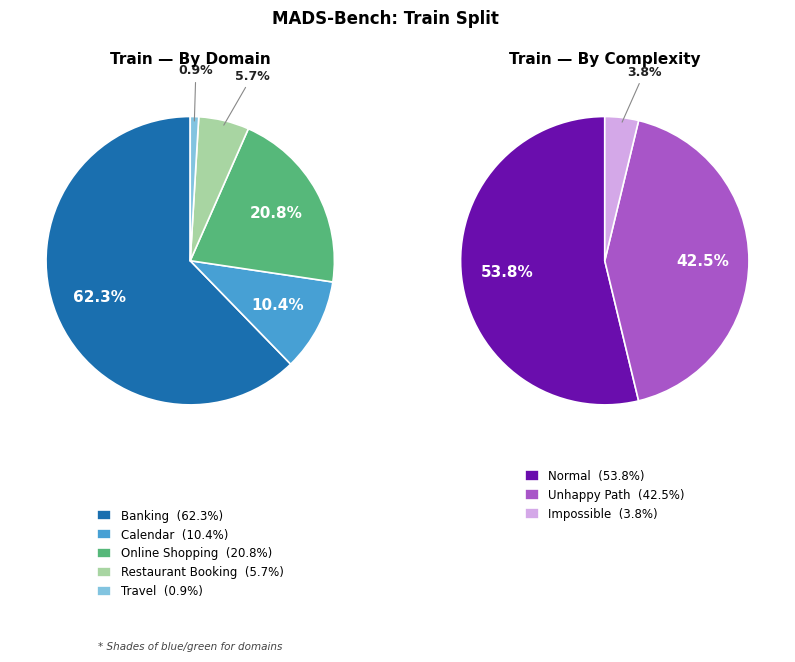}
    \caption{MADS-Bench training split distribution by domain (left) and scenario complexity (right). Banking dominates the training set (62.3\%), while unhappy-path scenarios account for 42.5\% of training conversations, reflecting the dataset's emphasis on non-cooperative user behaviors.}
    \label{fig:train_distribution}
\end{figure}

\begin{figure}[t]
    \centering
    \vspace{-1cm}
    \includegraphics[width=1\linewidth]{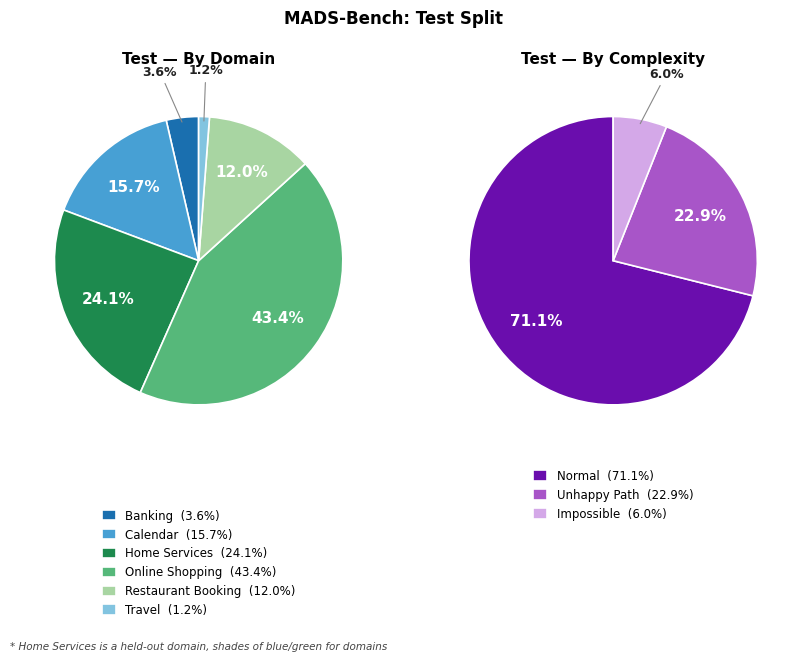}
    \caption{MADS-Bench test split distribution by domain (left) and scenario complexity (right). Home Services appears exclusively in the test set, serving as a held-out domain for cross-domain generalization evaluation. The test set is predominantly composed of normal scenarios (71.1\%), with unhappy-path and impossible scenarios comprising the remaining 28.9\%.}
    \label{fig:test_distribution}
\end{figure}

MADS-Bench comprises 189 validated conversations spanning 1,095 turns across 87 unique scenarios in six task domains: Banking, Calendar, Home Services, Online Shopping, Restaurant Booking, and Travel. A summary of the dataset statistics is provided in Table~\ref{tab:dataset_stats}. The dataset comprises 106 training conversations and 83 test conversations. As shown in figure~\ref{fig:train_distribution}, \ref{fig:test_distribution}, the Home Services domain is held out entirely from training and was used for evaluating the cross-domain generalization across the different stages of fine-tuning as described in section \ref{sec:experiments}. Scenario complexity is categorized into Normal, Unhappy path, and Impossible trajectories. The distribution of the scenario types across the training and test split is illustrated in Figure \ref{fig:train_distribution}, \ref{fig:test_distribution}. The training consists of a higher proportion of unhappy path scenarios, reflecting the model to be trained on unhappy paths and evaluating the model's robustness.

\subsection{Evaluation}

Inspired from Pre-Act \cite{rawat2025preact}, we evaluated the performance of models across two levels of evaluation. First, we compute the turn level evaluation (L-1), in which each conversation turn is evaluated against the ground-truth. Further, we evaluated the agents trajectory level success rate (L-2), assessing multi-turn reasoning and planning across the entire conversation trajectory. 

\subsubsection{Turn Level Evaluation (L-1)}

In turn level evaluation, we compare the model's performance for each conversation turn against the ground-truth response. For inference on the test split, we used the following two methods Pre-Act \cite{rawat2025preact} and ReAct \cite{yao2023react}. For each user query, the model must decide between two actions either to make a tool call and generate response based on the tool's response, or to have a conversation with the user to clarify the user query. We compute the following metrics: 

\textbf{Action Classification} We measure recall for the binary action decision (tool call vs. answer), which captures the model's ability to correctly identify the correct action to perform given the user query and conversation context. 

\textbf{Tool Call Accuracy} When the ground-truth action requires a tool call, we evaluate the tool invocation accuracy, whether the right tool was invoked, and whether the tool parameters matched, as well as whether all the required parameters were provided with the correct parameter values retrieved from the conversation context. 

\textbf{Conversation Quality} For turns where the ground-truth action is a direct user-facing response, we compute semantic similarity between the predicted and reference utterances using \texttt{sentence-transformers/all-mpnet-base-v2}\footnote{\url{https://huggingface.co/sentence-transformers/all-mpnet-base-v2}}. Cosine similarity between embeddings captures semantic alignment beyond surface-level token overlap.

\subsubsection{Trajectory Level Evaluation (L-2)}

We evaluate the models at the trajectory level by scoring whether a model can sustain correct multi-turn decision-making and action sequence over an entire conversation, rather than matching isolated turns. Concretely, we run the model on the held-out MADS test set and obtain a sequence of predicted assistant outputs, each containing a structured \texttt{<action>} block. We then group predictions into full conversations by \texttt{scenario\_id} and conversation boundaries, and score each conversation in temporal order using a strict ``stop-on-first-error'' criterion. A turn is marked correct if the predicted action type matches the reference (i.e., \texttt{tool} vs.\ \texttt{say}); for tool actions, the predicted tool name must match the reference tool name under case-insensitive exact matching with a small edit-distance tolerance. If a turn is incorrect, scoring terminates immediately for that conversation. The conversation score is defined as the fraction of turns that are correct up to the first error, i.e., $\texttt{correct}/\texttt{attempted}$ with attempted equal to the index of the first incorrect turn. The reported L2 success rate is the mean conversation score, which penalizes early failures and rewards agents that maintain consistent grounded behavior over longer turn conversations.

\section{Experiments}
\label{sec:experiments}
In this section, we demonstrate the effectiveness of MADS-Bench through a series of evaluations across different training stages. 
We frame the task of agentic dialogue output as a sequential decision-making problem where, at each step, the model receives the conversation history and must produce a single structured output containing reasoning (\texttt{<think>}), a plan (\texttt{<plan>}), and an action (\texttt{<action>}). Actions are either tool invocations or user-facing utterances. We adopted a two-stage training pipeline consisting of supervised fine-tuning (SFT) followed by reinforcement learning with verifiable rewards (RLVR) inspired by \cite{qian2025toolrl, rawat2025actionsteachthinkreasoningaction}.

\subsection{Training Setup}

\textbf{Supervised Fine-Tuning (SFT)} During the SFT stage, models are trained on the full MADS-Bench training split, consisting of 106 conversation trajectories, using a learning rate of $2\times10^{-4}$, batch size 4, with gradient accumulation over 4 steps. The objective is next-token prediction over structured outputs containing reasoning (\texttt{<think>}), planning (\texttt{<plan>}), and action tokens (\texttt{<action>}).

\textbf{Reinforcement Learning with Verifiable Rewards (RLVR)} RLVR training is performed on a subset of 50 sampled conversations from the training set. The RL stage uses a learning rate of $1\times10^{-5}$, batch size 1 with gradient accumulation of 4 steps, and is trained for 5 epoch. During training, the model generates multiple candidate responses ($K=8$ generations) for each rollout. We adopt the Group Relative Policy Optimization (GRPO) objective, following prior work \cite{qian2025toolrl}, which reports stronger performance compared to PPO in ablation studies.

\subsection{Supervised Fine-Tuning}

While supervised fine-tuning (SFT) substantially improved the model’s ability to follow the required format structure and increased tool-use frequency, it did not eliminate several critical failure modes that emerged repeatedly during trajectory-level evaluation.
Most notably, the assistant often:

\begin{itemize}
    \item Hallucinated task completion without invoking any tools. 

    \item Generated ungrounded or fabricated tool arguments, including nonexistent IDs and incorrect temporal values. 

    \item Ignored or contradicted tool outputs, e.g., promising a refund when the tool explicitly returned a zero-refund result.
\end{itemize}

These errors are not rare formatting mistakes but fundamental reasoning failures that compromise safety, robustness, and correctness in agentic systems. Crucially, they arise precisely in situations where the environment provides verifiable signals, tool schemas, argument types, tool outputs, and scenario constraints that a correct agent should incorporate. Therefore, we perform reinforcement learning post-training within our training pipeline to optimize the model for action correctness, accurate tool invoking and the semantic similarity of user-facing responses using signals that the environment can objectively verify. 

\subsection{Reinforcement Learning with Verifiable Rewards (RLVR)}

RLVR shifts the training objective from imitating demonstrations to behaving correctly according to tool outputs, argument validity, and action criteria.

To incorporate RL into our agent training pipeline, we adopt a GRPO policy gradient, but adapted to the design constraints of our Pre-Act simulation conversation. RLVR is applied on top of the SFT-trained model and operates by generating full rollouts, rewarding each assistant action against verifiable conversation state, and using the reward signals to update the policy toward action grounded and tool-dependent behavior.

Our RLVR setup uses three verifiable reward components that directly reflect the reward functions implemented in our code: (1) an action-type reward, (2) a tool-match reward, and (3) a final-answer F1 reward. Each component is computed deterministically from the model's generated \texttt{<action>} block and a gold reference action extracted using the same parsing logic. No heuristic scoring or human annotation is used.

\begin{itemize}
    \item \textbf{Action Reward} This reward measures whether the assistant predicted the correct high-level action type (\texttt{tool} vs.~\texttt{say}). The implementation compares the predicted action type with the gold action type extracted from the reference output. 

    \item \textbf{Tool Accuracy Reward} This reward is computed only when both the gold and predicted actions are tool invocations.  
It checks two conditions: (a) whether the predicted tool name matches the gold tool name, and (b) whether the predicted parameter dictionary exactly matches the gold parameters.

    \item \textbf{Semantic Reward} This reward is evaluated only when both the predicted and gold actions are of type \texttt{say}.  
Instead of relying on token-level overlap, we compute the reward using a sentence-transformer model that outputs a semantic similarity score between the predicted natural language response and the gold reference response. The predicted and gold utterances are embedded into a shared vector space, and their cosine similarity is used as the reward value. Formally:
\[
R_{\text{sem}} = \cos\bigl(\mathrm{embed}(\text{predicted}),\ \mathrm{embed}(\text{gold})\bigr).
\]
If either action is not a \texttt{say} action, the reward defaults to \textbf{0.0}.  
Since cosine similarity is bounded, we have $R_{\text{sem}} \in [0, 1]$.

\item 

\textbf{Total Reward} For each sample, the final reward passed to the RL optimizer is the sum of all three components: $R_{\text{total}} = R_{\text{act}} + R_{\text{tool}} + R_{\text{sem}}$. Given the individual ranges, $R_{\text{total}} \in [0, 3]$.

\end{itemize}

This formulation ensures that rewards remain bounded, sparse, and fully verifiable, aligning with the design principles of RL with verifiable signals.

% \begin{figure}[t]
%     \centering
%     \vspace{-1.2cm}
%     \includegraphics[width=0.4\textwidth,keepaspectratio]{SFT+RLVR.pdf}
%     \vspace{-.2cm}
%     \caption{Training Approach}
%     \label{fig:training_approach}
% \end{figure}

\subsection{Cross-Dataset Generalization: APIGen Baseline}

To evaluate whether MADS-Bench provides a training signal beyond what is available in existing function-calling datasets, we train a baseline model exclusively on data converted from APIGen MT~\cite{liu2024apigen} and evaluate it on our held-out MADS-Bench test set.

\textbf{Data Conversion.} APIGen-MT conversations use \texttt{human}, \texttt{gpt}, \texttt{function\_call}, and \texttt{observation} turns without explicit reasoning or 
planning traces. We convert 198 APIGen-MT conversations into our format via an automated pipeline: tool definitions are parsed from the dataset's JSON schema, turn roles are remapped to our message structure, and \texttt{<think>} and \texttt{<plan>} blocks are synthesized for each assistant turn using GPT-4o-mini conditioned on the conversation history and the original assistant response. Converted conversations undergo the same multi-stage validation pipeline as MADS-Bench, yielding 750 training samples.

\textbf{Training.} We fine-tune Qwen2.5-7B-Instruct exclusively on the converted APIGen data, using the same QLoRA configuration as described, allowing a direct comparison of the training data source while controlling for model architecture and fine-tuning procedure.

\textbf{Evaluation.} The APIGen-finetuned model is evaluated on our MADS-Bench test set at the L1 level, using the same metrics and evaluation infrastructure as all other models.

\section{Results and Analysis}

\subsection{L-1 Evaluation}

We evaluate turn-level behavior using structured action metrics, including action recall, tool-name F1, parameter match accuracy, final-answer F1, and semantic similarity. L1 evaluation isolates individual decision points and measures whether the model selects the correct action type, invokes the appropriate tool with correct arguments, and produces grounded user-facing responses. We analyze two orthogonal factors: (i) Inference time reasoning prompts (Pre-Act vs.\ ReAct) Table (\ref{tab:l1-supervision}) and (ii) Impact of RLVR (Table \ref{tab:l1-stages}).

\begin{table*}[t]
\centering
\caption{L1 (turn-level) evaluation on the MADS test set comparing \textbf{supervision sources}. \textbf{Approach} denotes the inference-time prompting format (Pre-Act vs.\ ReAct). \textbf{Supervision} indicates the dataset used for SFT (APIGen vs.\ MADS). All models are trained with a Pre-Act system prompt; ReAct is applied only at inference time.}
\small
\renewcommand{\arraystretch}{1.05}
\setlength{\tabcolsep}{4pt}
\begin{tabular*}{\textwidth}{@{\extracolsep{\fill}} lllccccc}
\toprule
\textbf{Model} & \textbf{Approach} & \textbf{Dataset} & \textbf{Action Recall} & \textbf{Tool F1} & \textbf{Param Match} & \textbf{Response F1} & \textbf{Response Sim.} \\
\midrule

\multirow{6}{*}{\textit{Qwen2.5-7B}}
& Pre-Act & Base        & 0.22 & 0.38 & 0.38 & 0.07 & 0.14 \\
& Pre-Act & SFT-APIGen  & 0.32 & 0.50 & 0.38 & 0.1 & 0.15 \\
& Pre-Act & SFT    & \textbf{0.60} & \textbf{0.70} & \textbf{0.44} & \textbf{0.25} & \textbf{0.51} \\
\cmidrule(lr){2-8}
& ReAct  & Base        & 0.51 & 0.68 & 0.43 & 0.12 & 0.29 \\
& ReAct  & SFT-APIGen  & 0.42 & 0.51 & 0.43 & 0.17 & 0.21 \\
& ReAct  & SFT    & 0.54 & 0.54 & 0.44 & 0.15 & 0.35 \\

\midrule

\multirow{6}{*}{\textit{Qwen2.5-32B}}
& Pre-Act & Base        & 0.35 & 0.64 & 0.35 & 0.11 & 0.22 \\
& Pre-Act & SFT-APIGen  & 0.46 & 0.67 & 0.63 & 0.24 & 0.49 \\
& Pre-Act & SFT    & \textbf{0.66} & \textbf{0.70} & \textbf{0.66} & \textbf{0.32} & \textbf{0.57} \\
\cmidrule(lr){2-8}
& ReAct  & Base        & 0.35 & 0.64 & 0.35 & 0.08 & 0.20 \\
& ReAct  & SFT-APIGen  & 0.4 & 0.68 & 0.66 & 0.33 & 0.46 \\
& ReAct  & SFT    & 0.64 & 0.62 & 0.34 & 0.34 & 0.60 \\

\bottomrule
\end{tabular*}
\label{tab:l1-supervision}
\end{table*}

\begin{table*}[t]
\centering
\caption{L1 (turn-level) evaluation on the MADS test set comparing \textbf{training stages on MADS}. Stage denotes the optimization stage (SFT $\rightarrow$ SFT+RLVR). Approach denotes the inference-time prompting format (Pre-Act vs.\ ReAct). All models are trained with a Pre-Act system prompt; ReAct is applied only at inference time.}
\small
\renewcommand{\arraystretch}{1.05}
\setlength{\tabcolsep}{4pt}
\begin{tabular*}{\textwidth}{@{\extracolsep{\fill}} lllccccc}
\toprule
\textbf{Model} & \textbf{Approach} & \textbf{Stage} & \textbf{Action Recall} & \textbf{Tool F1} & \textbf{Param Match} & \textbf{Response F1} & \textbf{Response Sim.} \\
\midrule

\multirow{4}{*}{\textit{Qwen2.5-7B}}
& Pre-Act & SFT  & 0.60 & 0.70 & 0.44 & \textbf{0.25} & \textbf{0.51} \\
& Pre-Act & SFT+RLVR   & \textbf{0.67} & 0.77 & \textbf{0.51} & 0.21 & 0.46 \\
\cmidrule(lr){2-8}
& ReAct  & SFT  & 0.54 & 0.54 & 0.44 & 0.15 & 0.35 \\
& ReAct  & SFT+RLVR   & 0.67 & \textbf{0.80} & 0.48 & 0.23 & 0.50 \\

\midrule

\multirow{4}{*}{\textit{Qwen2.5-32B}}
& Pre-Act & SFT  & 0.66 & 0.70 & 0.66 & 0.32 & 0.57 \\
& Pre-Act & SFT+RLVR   & \textbf{0.70} & \textbf{0.82} & \textbf{0.67} & 0.32 & 0.57 \\
\cmidrule(lr){2-8}
& ReAct  & SFT  & 0.64 & 0.62 & 0.34 & \textbf{0.34} & 0.60 \\
& ReAct  & SFT+RLVR   & 0.67 & 0.82 & 0.42 & 0.32 & \textbf{0.60} \\

\bottomrule
\end{tabular*}
\label{tab:l1-stages}
\end{table*}

\textbf{Inference-Time Reasoning Prompting (Pre-Act vs.\ ReAct).}
We further analyze the interaction between training supervision and prompting strategy. Pre-Act prompting generally yields higher action recall under MADS supervision, particularly for smaller models, suggesting that structured planning during training aligns naturally with Pre-Act inference.

However, ReAct prompting exhibits competitive or stronger tool F1 in some settings (e.g., Qwen-7B SFT+RLVR), indicating that step-wise reasoning may better leverage RL-refined policies. These results suggest that prompting strategy interacts with supervision type and model capacity, and that RLVR improves robustness under both prompting paradigms.

%\FloatBarrier

% \FloatBarrier
\textbf{Effect of RLVR}
We next examine the impact of RLVR refinement on models that have already been fine-tuned with MADS-bench supervision  referred in table \ref{tab:l1-stages}. RLVR consistently improves structured grounding metrics across both model sizes, with particularly strong gains in tool-level correctness. RLVR improves verifiable components of decision-making, particularly tool-name selection and parameter matching, more reliably than semantic answer quality. Larger models benefit more consistently from reward signals, suggesting the emergent capacity to internalize verifiable reward structures. Taken together, the L1 results indicate that while supervised learning establishes basic formatting and tool-use capability, RLVR is necessary to reduce hallucinated actions and fabricated arguments.

\subsection{L-2 Evaluation}

We evaluate models at the trajectory level, measuring whether they can sustain correct multi-turn decision-making throughout the conversation. Unlike L1, which isolates individual turns, L2 captures the compounding effects of early errors, tool misuse, and recovery failures. We report the mean conversation success rate on the held-out MADS test set.

\textbf{Overall Task Success Rate across Training stages} 

Table \ref{tab:l2-success} summarizes the overall trajectory-level task success rate across the model size and training stage. Across models, the RLVR stage consistently improves performance over the SFT stage, indicating improved prediction of the correct set of actions and invocation of the appropriate tool throughout the entire conversation trajectory. The results suggest that while SFT is important for guiding the model toward a structured action pattern, RLVR plays a key role in improving overall correctness across the entire conversation trajectory. 

For Qwen2.5-32B, SFT already achieves a stronger baseline success rate of 44.08\%, reflecting the larger model’s improved capacity for structured reasoning and action sequence. RLVR further increases performance to 53.96\%, yielding a consistent improvement. The smaller relative gain suggests that larger models internalize structured supervision more effectively during SFT, leaving less headroom for reinforcement refinement. Nonetheless, RLVR still improves long-horizon action sequence and reduces failure cascades.

\begin{table}[t]
\centering
\caption{L2 (trajectory-level) task success on the MADS-bench test set. Training Stage denotes optimization on MADS (SFT vs.\ RLVR).}
\small
\renewcommand{\arraystretch}{1.05}
\setlength{\tabcolsep}{6pt}
\begin{tabular}{llc}
\toprule
\textbf{Model} & \textbf{Training Stage} & \textbf{Task Success(\%)} \\
\midrule

\multirow{2}{*}{\textit{Qwen2.5-7B}}
& SFT  & 24.49  \\
& SFT+RLVR   & \textbf{39.78} \\

\midrule

\multirow{2}{*}{\textit{Qwen2.5-32B}}
& SFT  & 44.08  \\
& SFT+RLVR   & \textbf{53.96} \\

\bottomrule
\end{tabular}
\label{tab:l2-success}
\end{table}

%\FloatBarrier

\textbf{Cross-Domain Generalization} 

To understand the effect of domain supervision, Table \ref{tab:l2_domain} compares the performance of models trained on APIGen and MADS-Bench across domains. For the APIGen models trained on the retail and airline domains, and the MADS-Bench models trained across the five domains in the training set. We kept Home Services as the held-out domain to compare the performance of different training stages in improving the success rate across held-out domains. Interestingly, we observe that APIGen-trained models occasionally achieve higher trajectory success than MADS-trained models under supervised fine-tuning. This behavior arises because APIGen primarily contains cooperative task trajectories with deterministic tool execution patterns, which encourage stable but shallow decision policies. In contrast, MADS-Bench intentionally introduces adversarial conversational dynamics. While these scenarios improve model robustness during the RLVR stage of post-training, they don't significantly improve performance alone in the SFT stage. The RLVR stage consistently improves the performance across all the domains, importantly the same trend is seen across both the model sizes in the held out home services domain indicating RLVR stage improves the cross-domain generalization when applied as an additional step after supervised post training.

\begin{table}[t]
\centering
\caption{L2 task success rate (\%) by domain. Models are evaluated on the
MADS-Bench test set. APIGen models are trained with SFT only. MADS-Bench
models are trained with SFT and further refined with RLVR. The Home Services domain was excluded from training to evaluate cross-domain
generalization. Travel contains a single test instance.}
\small
\renewcommand{\arraystretch}{1.05}
\setlength{\tabcolsep}{0.2pt}
\begin{tabular}{lccccc}
\toprule
\textbf{Domain} & \textbf{APIGen} & \multicolumn{4}{c}{\textbf{MADS-Bench}} \\
                & \textbf{SFT} & \multicolumn{2}{c}{\textbf{SFT}} & \multicolumn{2}{c}{\textbf{SFT+RLVR}} \\
                & 7B & 7B & 32B & 7B & 32B \\
\midrule
Banking            & 33.33 & 44.44 & 16.67 & 33.33 & \textbf{66.67} \\
Calendar           & 45.51 & 35.00 & 46.15 & 57.44 & \textbf{69.23} \\
Home Services (Held Out) & 40.92 & 28.87 & 49.95 & 42.50 & \textbf{53.87} \\
Online Shopping    & 37.64 & 19.44 & 39.17 & 37.64 & \textbf{43.10} \\
Restaurant Booking & 0.00 & 16.67 & 50.00 & 25.00 & \textbf{65.00} \\
Travel Assistance  & 0.00 & 0.00  & 100.00 & 0.00 & \textbf{100.00} \\
\bottomrule
\end{tabular}
\label{tab:l2_domain}
\end{table}

\textbf{Scenario Complexity Analysis} 

% rerference table
% rewrite this pg
Table~\ref{tab:l2_complexity} shows task success rates broken down by task complexity, i.e., whether it can be completed ``normally'' (without problems), via an ``unhappy path'', or encounters an ``impossible'' scenario.
In cooperative normal scenarios, models trained by RLVR achieve the strongest performance, suggesting that RLVR helps follow the correct sequence of steps and reduces failure cases. Using SFT only, APIGen training is actually better than MADS-Bench training, given that the dataset is focused on ``normal'' conversations.
However, RLVR improves both the ``normal'' and ``unhappy path'' cases dramatically, and outperforms the APIgen baseline.
The RLVR improvement is largest for the small model, but still substantial for the large (32B) model.
% AS: the below is not true, AGPIgen training beats MADS-Bench even for unhappy paths.
%The benefit of SFT using MADS-Bench data becomes in unhappy path scenarios, which include adversity injected behaviors.
%Under these scenarios, RLVR provide the largest benefit across both the models indicating that RL post training is important in improving model robustness across unhappy scenarios.
%This also shows that the involvement of unhappy paths in training across various stages of post training helps in %improving the performance across both small and large agentic models. 
We also find that RLVR degrades the performance for ``impossible'' conversations.  This could simply be due to a lack of training data, since our limited RL fine-tuning set has only five instances of impossible cases.

In terms of model size, we note that the 32B model performs overall better than the 7B version, but that the advantage is especially large for the harder conversations (unhappy and impossible). It makes sense that larger models have better reasoning ability, allowing them to deal better with more complex tasks.

\begin{table}[t]
\centering
\caption{Task success (\%) by scenario complexity for different model configurations (training data and objective). Unhappy-path
scenarios involve non-cooperative user behaviors, such as goal
changes and tool failures. Impossible scenarios require the
agent to recognize task infeasibility and respond gracefully.
APIGen models are trained only with SFT. MADS models are trained
with SFT and further refined with RLVR. All models are evaluated
on the MADS-Bench test set.}
\small
\renewcommand{\arraystretch}{1.05}
\setlength{\tabcolsep}{2pt}
\begin{tabular}{lccccc}
\toprule
\textbf{Complexity} & \textbf{APIGen} & \multicolumn{4}{c}{\textbf{MADS-Bench}} \\
                    & \textbf{SFT}   & \multicolumn{2}{c}{\textbf{SFT}} & \multicolumn{2}{c}{\textbf{SFT+RLVR}} \\
                    & 7B    & 7B & 32B & 7B & 32B \\
\midrule
Normal       & 35.82 & 23.57 & 38.06 & 40.28          & \textbf{48.94}          \\
Unhappy Path & 29.04 & 23.25 & 57.72 & 39.04          & \textbf{68.86} \\
Impossible   & 40.0 & 40.00 & \textbf{63.33} & 36.67          & 56.67 \\
\bottomrule
\end{tabular}
\label{tab:l2_complexity}
\end{table}

% \begin{figure}
%     \centering
%     \includegraphics[width=1\linewidth]{tool_call_vs_conv_len.png}
%     \caption{Number of tool calls versus conversation length}
%     \label{fig:placeholder}
% \end{figure}

\section{Conclusions}

%Authors must proofread their PDF file prior to submission, to ensure it is correct. Do not rely on proofreading the \LaTeX\xspace source or Word document. \textbf{Please proofread the PDF file before it is submitted.}

Real-world task-oriented conversational systems must operate under conditions that diverge substantially from the cooperative settings typically represented in existing function-calling benchmarks. Users frequently provide incomplete information, revise goals during interaction, or issue requests that cannot be satisfied due to system constraints or policy limitations. Despite the rapid progress of LLM-based agents, the lack of training resources that systematically capture such failure-prone conversational dynamics remains a key bottleneck for building robust dialog managers.

In this work, we introduced MADS, a multi-agent dialog simulation pipeline designed to generate structured multi-turn conversations that include reasoning, planning, and tool-invocation supervision. By orchestrating interactions between simulated user, assistant, and tool agents, the pipeline enables controllable injection of both conversational adversities (e.g., partial information, goal revisions) and execution-level failures (e.g., backend errors, policy constraints). Using this simulator, we constructed MADS-Bench, a multi-domain benchmark spanning six task-oriented domains, 72 tools, and diverse user personas, with a large proportion of scenarios representing unhappy-path or impossible interactions.

Empirical results demonstrate that training on MADS-Bench improves the robustness of tool-augmented agents beyond that achievable with cooperative datasets alone. While supervised fine-tuning establishes structural competence for reasoning and tool usage, reinforcement learning with verifiable rewards (RLVR) significantly improves action correctness and trajectory-level task success across both small and large models. These findings suggest that verifiable reward signals derived from tool-grounded environments are critical for stabilizing long-horizon decision making in conversational agents.

Future work may extend this pipeline along several directions. First, scaling the simulator to larger tool ecosystems and more diverse adversarial behaviors could further improve the coverage of realistic dialog conditions. Second, integrating real tool execution environments or sandboxed APIs may enable richer reward signals and more faithful agent evaluation. Finally, extending the simulator to multimodal or speech-to-speech systems would enable the study of robustness in next-generation conversational assistants operating in real world human–AI interaction scenarios.

Overall, MADS provides a scalable approach to generating structured training data that exposes conversational agents to realistic failure modes, thereby enabling the development of more robust and reliable task oriented conversational agents.

\clearpage

\section{Acknowledgments}
We appreciate the efforts of the developers of the Qwen 2.5 model family and the maintainers of the open‑source libraries that supported our pipeline.

We are also appreciative to Interspeech for providing a platform to share our work with the speech and language processing community.

%Thanks to Work on this! 

{%\color{blue}
%\section{Generative AI Use Disclosure}
%Generative AI tools were used in a limited capacity to assist with language editing, clarity improvement, and formatting of the manuscript. All experimental design, dataset construction, analysis, and scientific claims were developed by the authors. The authors take full responsibility for the content of the paper and have verified all results and conclusions. No generative AI system was used to produce experimental results, generate data, or make scientific decisions.
%}

%\section{Reference Format}
%
%{\color{blue}It is ISCA policy that papers submitted to Interspeech should refer to peer-reviewed publications. References to non-peer-reviewed publications (including public repositories such as arXiv, Preprints, and HAL, software, and personal communications) should only be made if there is no peer-reviewed publication available, and should be kept to a minimum.

%References should be in standard IEEE format (follows the IEEEtran format), numbered in order of appearance, for example \cite{Davis80-COP} is cited before \cite{Rabiner89-ATO}. For longer works such as books, provide a single entry for the complete work in the References, then cite specific pages \cite[pp.\ 417--422]{Hastie09-TEO} or a chapter \cite[Chapter 2]{Hastie09-TEO}. Multiple references may be cited in a list \cite{Smith22-XXX, Jones22-XXX}.}

\bibliographystyle{IEEEtran}
\bibliography{mybib}

\end{document}

%%% Local Variables:
%%% mode: LaTeX
%%% TeX-master: t
%%% End: